\documentclass{article}

\usepackage[preprint]{neurips_2024}
\usepackage{natbib}
\setcitestyle{numbers,square}

\usepackage[utf8]{inputenc} 
\usepackage[T1]{fontenc}    
\usepackage{hyperref}       
\usepackage{url}            
\usepackage{booktabs}       
\usepackage{amsfonts}       
\usepackage{nicefrac}       
\usepackage{microtype}      
\usepackage{xcolor}         
\usepackage{graphicx}       
\usepackage{float}          
\usepackage{subfigure}      
\usepackage{algorithm}
\usepackage{algorithmic}
\usepackage{multirow}
\usepackage{wrapfig}

\title{Transferable text data distillation by trajectory matching}

\author{%
Rong Yao \quad Hailin Hu \quad Yifei Fu \quad Hanting Chen \quad Wenyi Fang \\
\quad  \textbf{Fanyi Du} \quad \textbf{Kai Han} \quad \textbf{Yunhe Wang}\\
Huawei Noah’s Ark Lab\\
\texttt{\{yaorong,hailin.hu,fuyifei1,chenhanting,fangwenyi1, }\\
\texttt{dufanyi,kai.han,yunhe.wang\}@huawei.com}\\
}

\begin{document}

\maketitle


\begin{abstract}
In the realm of large language model (LLM),  as the size of large models increases, it also brings higher training costs. There is a urgent need to minimize the data size in LLM training. Compared with data selection method, the data distillation method aims to synthesize a small number of data samples to achieve the training effect of the full data set and has better flexibility. Despite its successes in computer vision, the discreteness of text data has hitherto stymied its exploration in natural language processing (NLP). In this work, we proposed a method that involves learning pseudo prompt data based on trajectory matching and finding its nearest neighbor ID to achieve cross-architecture transfer. During the distillation process, we introduce a regularization loss to improve the robustness of our distilled data. To our best knowledge, this is the first data distillation work suitable for text generation tasks such as instruction tuning. Evaluations on two benchmarks, including ARC-Easy and MMLU instruction tuning datasets, established the superiority of our distillation approach over the SOTA data selection method LESS.  Furthermore, our method demonstrates a good transferability over LLM structures (i.e., OPT to Llama).
\end{abstract}

\section{Introduction}  \label{secintro}

In recent years, with the rapid development of large language models (LLMs)\cite{brown2020language,zhang2022opt,touvron2023llama, anil2023palm}, applications related to LLMs have become increasingly widespread. However, as language models grow larger, the required amount of training data also increases, leading to higher training costs. In companion with model size scaling, the quantity of required data is also piling up significantly, leading to significant computational burden and quality concern.

To reduce the training costs of LLM, some studies \cite{iter2021complementarity,grangier2022tradeoffs,bejan2023make,Zhou_2023_ICCV,xia2024less,xie2023data} have proposed a series of data selection methods that involve selecting a subset of text data to replace the entire text dataset. For example, DQ~\cite{Zhou_2023_ICCV} divides the original dataset into a set of non-overlapping bins, and then selects samples from all bins, which ensures the diversity of data selection. LESS~\cite{xia2024less} uses the correlation of the gradient features of the original and validation data as an influencing factor. The data selection method relies on the designed metrics to remove redundant data for that are useless for training. However, without modifying the data, the performance of purely selection-based method is bounded by the existence of highly-relevant data as a premise.

As another type of method, some studies from the field of computer vision \cite{sachdeva2023data, cazenavette2022dataset} have already demonstrated that data distillation is a more effective method for reducing dataset size than data selection. Data distillation aims to synthesize a small dataset to achieve the training effect of the full data set and can be divided into different classes based on their distillation targets, i.e., meta-model matching, gradient matching, trajectory matching and distribution matching \cite{sachdeva2023data}. Among these, trajectory matching has been proven to exhibit stable and excellent performance by overcoming the limitation including the short-range nature of single-step matching and instability of optimization.

However, the current mainstream of data distillation is established in the field computer vision, which takes the advantage of distinct class labels and continuous feature space \cite{sachdeva2023data} to generate task-specific pixel sets to concretely store the raw information for useful for training different models. In the training of LLM, however, the training objective is not a class label, which diversifies the optimization direction of a given sample. Moreover, each data point consists of a sequence of tokens that are discrete in the specific vocabulary list associated with each model, making it more difficult to transfer the generated samples across models. In this paper, we propose a text data distillation method, named neighbor-aware corpus distillation (NACD), to effectively improve the text data for better training performance. This framework is compatible with the current instruct tuning procedure and can generate useful but light-weight training data that absorbs the training trajectory from the large-scale dataset.

In particular, to avoid the intractable optimization of discrete token combination along the sequence, we propose to inject learnable tokens to each data point and leverage the diverge optimization language model itself to integrate the injected information. Moreover, to improve the robustness of the learned pseudo data, we design a neighbor-based regularizar to restrict the samples in feature spaces close to existing semantic tokens. Empirical studies show that our method can significantly improve the instruct tuning model of mainstream models such as OPT \cite{zhang2022opt} and Llama-2 \cite{touvron2023llama}. Moreoever, we show that even sophisticatedly designed gradient data selection method such as LESS \cite{xia2024less} can still be optimized on common benchmarks, opening up a new perspective for data compression regarding LLM training. Our main contributions are as follows:

\begin{itemize}
    \item We proposed a text data distillation method which combines prompt tuning and trajectory matching. The method of concatenating prompt data effectively harnesses the flexibility of data distillation techniques while preserving the information of the text data. The effectiveness of our approach has been validated on two datasets and multiple language models.
    \item We proposed a method that involves learning pseudo prompt data and finding its nearest neighbor ID to achieve cross-architecture transfer. To our knowledge, this is the first work that solving the text data distillation problem of text generation tasks in different language models.
    \item To enhance the robustness of the text data distillation process, we proposed a regularization loss based on more certain semantic token distributions, which improves the performance of text distillation tasks.
\end{itemize}

\section{Related works}
\subsection{Data distillation}
Data distillation (DD) aims at extracting knowledge from a large dataset and distilling it to a small synthetic dataset while the model is trained to achieve similar performance compared to the original dataset. Wang et al.~\cite{wang2020dataset} first introduced the idea of data distillation based on \textbf{meta learning}, where synthetic datasets are constructed by gradient descent, and the trained model has a low loss on the original dataset. Further, LD~\cite{bohdal2020flexible} distills labels instead of images, the method is flexible and effective. Zhao et al.~\cite{zhao2021dataset} proposed \textbf{gradient matching} for data distillation, making the gradient descent of the training process on the synthetic dataset similar to that on the original dataset. Differentiable Siamese Augmentation (DSA)~\cite{pmlr-v139-zhao21a} uses the same data augmentation for original and synthetic data to further improve distillation. Cazenavette et al. proposed MTT~\cite{cazenavette2022dataset} to directly match training trajectories between original and synthetic datasets. The \textbf{trajectory matching} approaches store the training trajectories of multiple expert networks on the original datasets, and the synthetic datasets are updated by matching the training trajectories. This method is effective in mitigating error accumulation. Considering the high computational cost associated with multiple optimizations required by the above methods, Zhao et al. proposed DM~\cite{Zhao_2023_WACV}, a method based on \textbf{distribution matching}, that uses single-step optimization to achieve data distillation. The method embeds original and synthetic data into the randomly sampled networks that learn the synthetic data by minimizing the distribution differences. CAFE~\cite{Wang_2022_CVPR} better characterizes original data distribution by aligning the layer-wise features between the original and synthetic datasets and introducing a bi-level optimization scheme.

\subsection{Data distillation in linguistic modality}
The discreteness of text data does not allow image data distillation methods to migrate well to text data.
Ilia et al.~\cite{Sucholutsky_2021} proposed to distill data and labels simultaneously, extending data distillation to text data by assigning soft labels to synthetic samples. Li et al.~\cite{li2021data} proposed a method to distill the text data into the embedding matrix, which solved the influence of the discreteness of text data on the distillation method. Maekawa et al.~\cite{maekawa-etal-2023-dataset} used distilled attention labels to extract knowledge from the original data to guide the optimization of the attention module. Sahni et al.~\cite{sahni2023exploring} similarly embedded text in continuous space and extended data distillation to multilingual data. DiLM~\cite{maekawa2024dilm} minimizes the gradient matching loss of synthetic samples by training a language model to bypass non-differentiable synthetic text. However, none of these methods can solve the related text-generation tasks.

\subsection{Data selection for task-specific fine-tuning}
Data selection is another solution to address the non-differentiability of text data by constructing small datasets through selection rather than generation. Iter et al.~\cite{iter2021complementarity} proposed to categorize the data into in-domain and out-of-domain during the fine-tuning phase and train BERT models as classifiers to select in-domain data for training. Grangier et al.~\cite{grangier2022tradeoffs} used three methods to select data: importance sampling, contrastive data selection, and influence functions. Bejan et al.~\cite{bejan2023make} used self-influence scores to capture outliers and automated curriculum learning (AutoCL) to replace fixed thresholds for data selection flexibility. DQ~\cite{Zhou_2023_ICCV} and LESS~\cite{xia2024less} perform coreset selection through the utilization of binning and correlation of the gradient features to avoid selecting a large number of samples in the high-density region of the dataset distribution.

\section{Methods}
\begin{figure}[ht] \label{mainfigure}
	\centering
	\begin{minipage}[c]{0.65\textwidth}
		\centering
		\includegraphics[width=\textwidth]{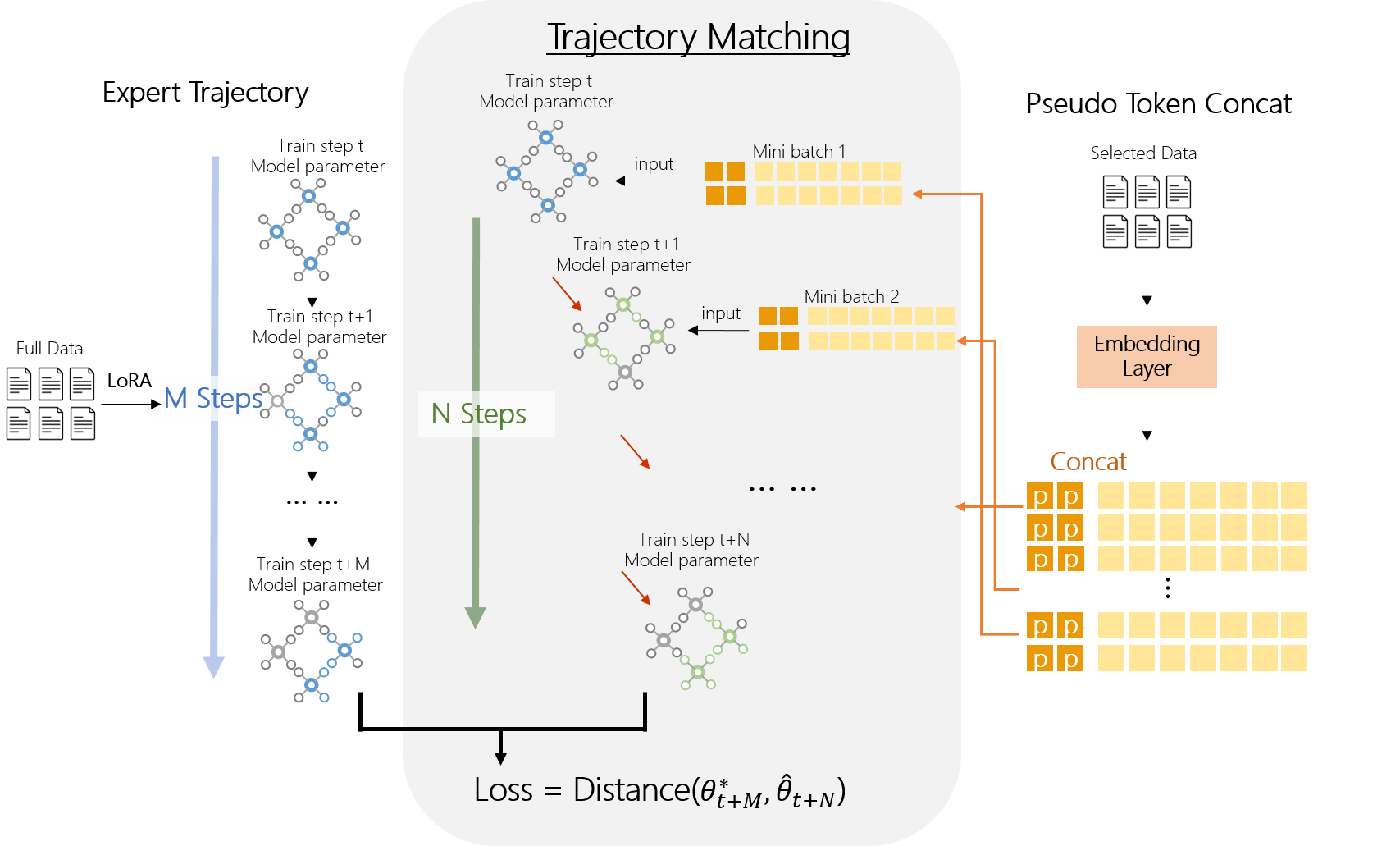}
        \centerline{(a)}
	\end{minipage} 
	\begin{minipage}[c]{0.3\textwidth}
		\centering
		\includegraphics[width=\textwidth]{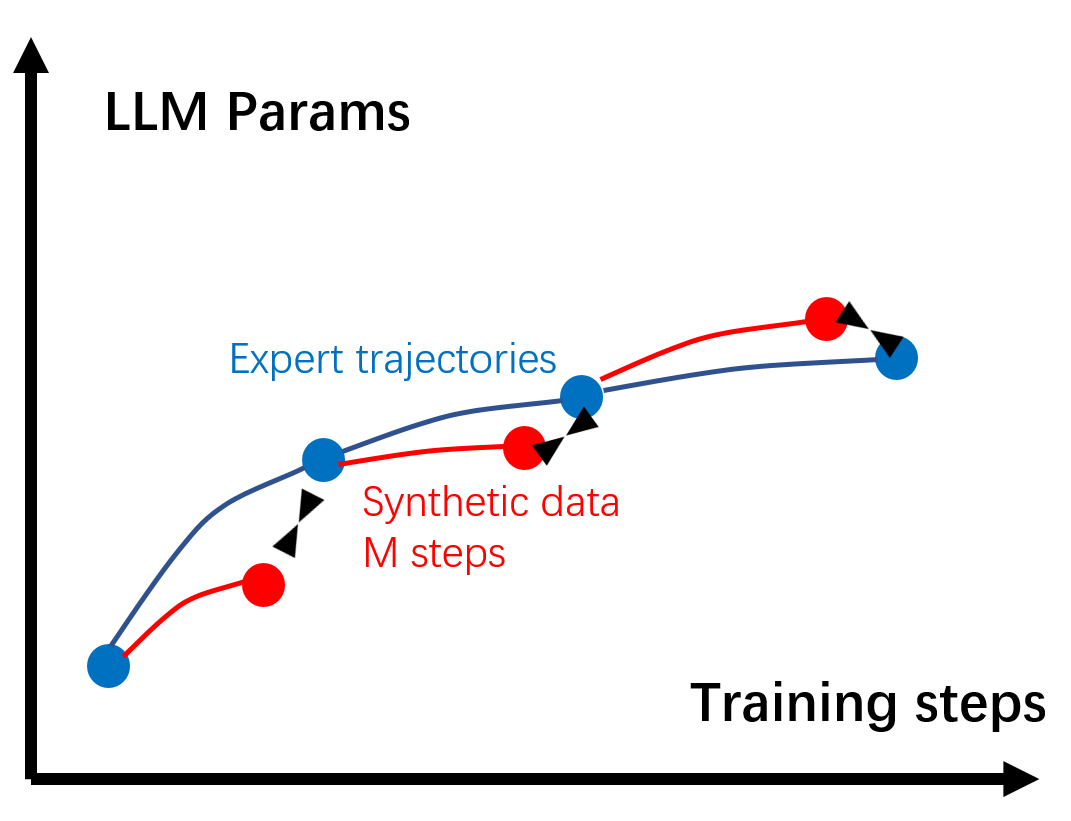}
        \centerline{(b)}
	\end{minipage}
    \caption{\textbf{(a)}: Illustration of our method. Step 1, we train the LLM with LoRA using full text data and saving all the intermediate parameters as expert trajectories. Step 2, given a small selection of text data $\mathcal{D}_{sel}$ (chosen by any data selection method), the synthesize prompt dataset $\mathcal{D}_{syn}$ is learned by trajectory matching aims to fit the learning trajectories of full text data. Step 3, after obtaining the prompt embedding, it is concatenated with the text data in the sequence dimension to accomplish instruction tuning. Step 4, evaluating the learned LLM in target data. \textbf{(b)}: A trajectory matching example. The blue trajectory represents parameter updates during training with full text data, while the red trajectory represents parameter updates during training with a subset of synthetic text data over N steps}
\end{figure}

In this section, we will describe our proposed text data distillation method. As mentioned earlier, data distillation methods aim to learn from a small amount of synthesized data to approximate results from the entire dataset. The discreteness of text data remains a pressing challenge. As illustrated in Figure {\ref{mainfigure}}, we achieve text data distillation by integrating data selection, prompt learning, and trajectory matching methods. The goal of our data distillation is to learn a small synthetic training text dataset $\mathcal{D}_{syn}$ such a model trained on this small dataset can achieve the training effect of the full data set $\mathcal{D}_{full}$. And the whole process can be divided to four steps. Specially, in Section 3.1 we discussed the expert trajectories extraction methods with full text data. In Section 3.2, we proposed the text data distillation method by trajectory matching. In Section 3.3, the instruction tuning and evaluation methods were discussed.

\subsection{Expert trajectories extraction with full text data}
Inspired by data distillation methods in the field of computer vision, especially trajectory matching methods, our approach is also based on performing long-range parameter trajectory matching between training of the distilled synthetic data and training on full data (Figure \ref{Fig.main2}). To accomplish trajectory matching, the first step involves extracting expert parameter trajectories from the language model. Similar to \cite{cazenavette2022dataset}, for language model, we need to first obtain a list of model parameters at different time points during the training process on the full dataset ${\{{\theta_{t}^{*}}\}}_0^T$, where $T$ denoted the number of total training steps. These long-range parameters represent the learning process of the model on the entire dataset. We take the parameters separated by a fixed step length as the start and end point of an expert parameter trajectory.

It should be noted that each independent training process will produce a parameter trajectory list ${\{{\theta_{t}^{*}}\}}_0^T$ and we save the snapshot parameters of each epoch in each training process. To increase the richness of the trajectory, the full dataset $\mathcal{D}_{full}$ will be used multiple times for training to generate a complete expert trajectory list, represented as $\{\tau_{i}^{*}\}$. In addition, the computation of the expert trajectories only requires the full dataset so that a pre-computation is allowed.

\subsection{Trajectory matching to distill prompt data} \label{sec32}
After obtaining expert trajectories, our objective is to distill the parameter update path of a small amount of synthetic text data to fit the full text data. Unlike the distillation process for synthetic data in computer vision, text data inherently possesses discrete properties. Particularly for text generation tasks, prior methods \cite{Sucholutsky_2021, li2021data, maekawa-etal-2023-dataset, sahni2023exploring} of learning full sentence representations are hard to optimize. Furthermore, it's inconvenient to transfer the learned synthetic data between model architectures. In contrast to learning representations from full data, we present a data distillation method based on Prompt learning. This approach involves learning synthetic data representations solely from prompts rather than full sentences. The main algorithm is shown as Algorithm \ref{alg:alg}.

\begin{figure}[htbp]
  \centering
  \includegraphics[width=0.9\textwidth]{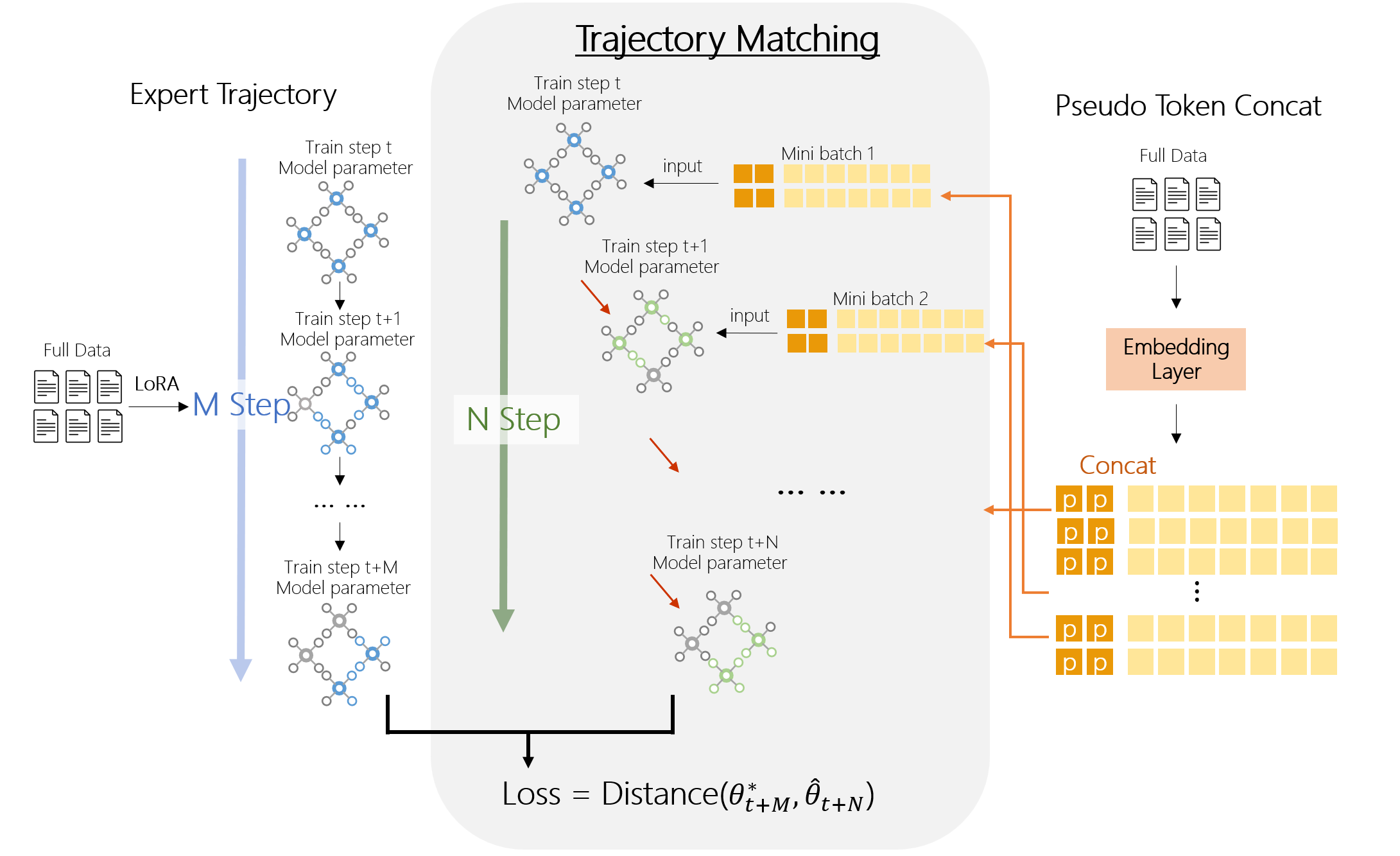} 
  \caption{Framework of our proposed text distillation process. We concatenate the learned pseudo token embeddings and the raw text data, and perform the LLM parameter trajectory matching by ensuring the student net has a similar model parameter updating to the experts.}
  \label{Fig.main2}
\end{figure}

\renewcommand{\algorithmicrequire}{ \textbf{Input:}} 
\renewcommand{\algorithmicensure}{ \textbf{Output:}} 
\begin{algorithm}
\caption{Text data distillation via prompt learning and trajectory matching}
\label{alg:alg}
\begin{algorithmic}[1] 
\REQUIRE $\{\tau_{i}^{*}\}$: set of expert trajectories learned on $\mathcal{D}_{full}$\\ 
\REQUIRE $\mathcal{D}_{sel}$: subset of $\mathcal{D}_{full}$\\ 
\REQUIRE $M$: step length of the target expert trajectory\\ 
\REQUIRE $N$: step length of the student model\\ 
\REQUIRE $S$: iteration steps of data distillation\\ 
\REQUIRE $k,d$: the prompt token length and the vocab embedding size of student model\\ 
\REQUIRE $\beta$: the weight factor of the regularization loss\\ 
\REQUIRE $T^m<T$: maximum start epoch\\ 
\STATE Initialize synthetic prompt data with average vocab embedding {$\mathcal{D}_{syn}\in \mathbb{R}^{(|\mathcal{D}_{sel}|,k,d)}$}  \\
\STATE Initialize learning rate $\alpha := \alpha_0$ \\
\FOR{$s \in [1,S]$}
\STATE sample an expert trajectory: $\tau^* \sim \{\tau_{i}^{*}\}$ with $\tau^*=\{{\theta_{t}^{*}}\}_0^T$ \\ 
\STATE sample a random start epoch $t<T^m$\\
\STATE initialize student net with expert parameters: $\hat{\theta}_{t}:=\theta_{t}^*$\\
\FOR{$n \in [1,N]$}
\STATE sample a mini-batch of prompt data and text data with the same indices: \\$p_{t+n}\sim\mathcal{D}_{syn},b_{t+n}\sim\mathcal{D}_{sel}$\\
\STATE update the student net parameters: $\hat{\theta}_{t+n+1}=\hat{\theta}_{t+n}-\alpha\nabla l(concat(p_{t+n}, b_{t+n}), \hat{\theta}_{t+n})$\\
\ENDFOR
\STATE Compute the distillation loss between ending student and expert parameters according to Eq. \ref{total-loss}: $\mathcal{L}=\mathcal{L}_{distill}+\beta\mathcal{L}_{reg}$\\
\STATE Update $\mathcal{D}_{syn}$ and $\alpha$ with $\mathcal{L}$
\ENDFOR
\ENSURE synthetic prompt data $\mathcal{D}_{syn}$; 
\end{algorithmic}
\end{algorithm}

We iteratively perform data distillation steps by trajectory matching to generate synthetic data $\mathcal{D}_{syn}$ through recurrent distillation. In order to enhance the transferability of learned synthetic data, our data distillation method, as a substitute for learning complete sentence embeddings, is based on text dataset obtained through data selection methods. After selecting a small text dataset $\mathcal{D}_{sel}$, We define $\mathcal{D}_{syn}\in \mathbb{R}^{(|\mathcal{D}_{sel}|,k,d)}$ based on $\mathcal{D}_{sel}$ to serve as its prompt token embedding, where $k$ denotes the prompt length and $d$ denotes the vocab embedding size.

At each distillation step, firstly, an expert parameter trajectory $\tau^*=\{{\theta_{t}^{*}}\}_0^T$ and a random start epoch $t$ are sampled. The student LLM $\hat{\theta}_{t}$ is initialized with ${\theta}_{t}^*$, then the student network will iteratively update N times to fit the expert trajectory. Taking the $n$th iteration as an example, in order to update the student network parameters, a mini-batch of data with the same index is simultaneously sampled from  $\mathcal{D}_{syn}$ and $\mathcal{D}_{sel}$ and used to calculate gradients.

\begin{equation}
\hat{\theta}_{t+n+1}=\hat{\theta}_{t+n}-\alpha\nabla l(concat(p_{t+n}, b_{t+n}), \hat{\theta}_{t+n}),
\end{equation}

where $p_{t+n}$ denotes the sampled prompt data, $b_{t+n}$ denotes the sampled text data, $concat(\cdot,\cdot)$ denotes concatenating two data embedding along the sequence dimension (i.e., the text data is passed to the embedding layer firstly), and $\alpha$ denotes a trainable learning rate. 

To update the student LLM parameters, a mini-batch of data with the same index is simultaneously sampled from datasets $\mathcal{D}_{syn}$ and $\mathcal{D}_{sel}$, and used to compute the gradients. After $N$ step gradient descent updates are performed, the final step's student network parameters are compared with target parameters. Similar to \cite{xia2024less}, we aim to learn from a small amount of synthetic data over N steps to fit the target parameters updated after $M$ steps on the full dataset (Figure \ref{Fig.main2}). The distillation loss function is defined as:
\begin{equation}
\mathcal{L}_{distill}=\frac{\Vert\hat{\theta}_{t+N}-\theta_{t+M}^*\Vert_2^2}{\Vert\hat{\theta}_{t}-\theta_{t+M}^*\Vert_2^2}
\end{equation}

In the distillation process, the loss function is normalized by the $l_2$ error between the target student net parameters $\theta_{t+M}^*$ and the N-step learned updated parameters $\hat{\theta}_{t+N}$. In particular, to improve the robustness of the learned prompt embeddings, we introduce a regularization loss to ensure that the learned representation is closer to a real text token:
\begin{equation}\label{reg_eq}
\mathcal{L}_{reg}=\frac{1}{N}\sum_{n=1}^{N}dis(p_{t+n},v_n),
\end{equation}

where $dis(\cdot,\cdot)$ denotes a distance calculation function, i.e., euclidean distance in our experiments, and $v_n$ represents the embedding that is closest to $p_{t+n}$ in the student net's embedding layer. In other words, we aim for the prompt embedding learned in the end to have the closest possible nearest neighbor token. Introducing a weight factor $\beta$, the total loss function for the distillation stage is defined as:
\begin{equation}\label{total-loss}
\mathcal{L}=\mathcal{L}_{distill}+\beta\mathcal{L}_{reg},
\end{equation}

And the distillation loss is minimized to update $\mathcal{D}_{syn}$ and $\alpha$. After multiple iterations, the learned prompt data representation $\mathcal{D}_{syn}$ is obtained, which can be used to concatenate in front of the selected dataset $\mathcal{D}_{sel}$, aiming to approximate or even surpass the performance of the full dataset when training a language model.

\subsection{Instruction tuning with concat data and evaluation}
To validate the effectiveness of learning synthetic data, we conducted experiments in the instruction tuning task by fine-tuning a Language Model with synthetic data. On one hand, for language models that have an embedding layer consistent with the distillation process, the prompt data obtained from distillation can be directly used as model input. All text data, after passing through the embedding layer to obtain sequence embeddings, can be concatenated directly with the corresponding prompt embedding and used as input for the language model. On the other hand, for language models with embedding layers that do not align during the distillation process, we propose a method to find the nearest token IDs corresponding to all prompt embeddings. These token IDs are converted into text and then new model token IDs, concatenated with the input token IDs of the new model, and used as input for the language model.

During the fine-tuning stage, the cross-entropy loss function is utilized for instruction tuning training. The LLM trained with synthetic data can then be used for evaluation.
\cite{cazenavette2022dataset}

\section{Experiments} \label{sec_exp}

\subsection{Dataset, model and experimental setup}  \label{sec_exp_set}
\textbf{Training and testing datasets:} To validate the effectiveness of the text data distillation method we propose, we chose the instruction tuning task for verification. Referring to \cite{xia2024less}, we selected four instruction tuning datasets—Flan v2 \cite{longpre2023flan}, CoT \cite{wei2022chain}, Dolly \cite{yuan2023evaluating} and Open Assistat 1 \cite{kopf2024openassistant}—as the target full text dataset for the distillation process. In order to thoroughly validate our method, the target dataset was set to "Flan v2 only" and "all four datasets" (referred to as $All^*$ in this paper) separately. The training datasets utilized in this study exhibit a high degree of diversity and are extensively employed in instruction tuning and enhancing the reasoning skills of language model. Moreover, we selected two multiple-choice question answering datasets MMLU \cite{hendrycks2020measuring} and ARC-Easy \cite{clark2018think} for testing. MMLU comprises 57 fields such as US history, mathematics, econometrics, etc., with each question containing four candidate answers. ARC-Easy is a dataset of grade-school level multiple-choice science questions, also presented in a multiple-choice format. These two datasets are frequently utilized to evaluate the reasoning capabilities of LLM.

\textbf{Models for data distillation and training:} 
In the data distillation phase, we chose OPT-1.3B \cite{zhang2022opt} as the model for expert trajectory extraction and trajectory matching. During the validation of distilled data, we respectively selected OPT-1.3B and Llama-2-7B \cite{touvron2023llama} as the models for instruction tuning and testing. OPT-1.3B serves as the original model in the distillation phase, sharing embedding space with the distilled data. Llama-2-7B differs from OPT-1.3B in its embedding layer, used to verify the cross-architecture effectiveness of our proposed method. Throughout all stages of data distillation and instruction tuning experiments, due to experimental cost constraints, we utilized LoRA \cite{hu2021lora} during all model training.

\textbf{Experimental setup:} 
As discussed in section \ref{sec32}, our approach first requires selecting a portion of data from the entire text dataset. In our experiments, we utilized the LESS \cite{xia2024less} based on Llama-2-7B to perform data selection, extracting 5\% of the data from the full text dataset firstly. And we found during the distillation stage experiments that initializing the synthetic prompt data with the mean of the OPT 1.3B vocab embedding leads to a more stable training process. In the distillation step, the step length of the student model $N$ is set to 6, the target expert trajectory step length is equal to an entire epoch, the iteration steps of distillation steps $S$ is set to 3000, the prompt token length $k$ is set to 2, the maximum start epoch is set to 2, and the weight factor of the regularization loss $\beta$ is set to 100. The batch sizes of LLM during trajectory extraction, data distillation and instruction tuning stages are set to 16, 8, and 8 respectively. In the test set metrics computation, when the base model is OPT-1.3B, with reference to \cite{zhang2022opt}, we compared the average likelihood of the candidate answers and calculated accuracy. When the base model is Llama-2-7B, with reference to \cite{xia2024less}, we compared the 5-shot accuracy.

\subsection{Main results} \label{mainresultsec}
\textbf{For the embedding layer consistency model:} As mentioned above, unlike traditional data distillation methods, our approach generates synthetic data in the form of prompt embedding data. To validate the performance of the learned prompt embedding data across various LLMs, we first chose the distillation stage model OPT-1.3B as the base model for instruction tuning. Since OPT-1.3B and synthetic prompt data share the embedding space, they can be directly concatenated along the sequence dimension for training OPT-1.3B.

We evaluated our method on MMLU and ARC-Easy datasets. As comparison methods, we also tested using the base model directly (denoted as "baseline"), training on the full dataset (denoted as "FULL"), and training on the raw data selected only (denoted as "LESS", our re-implementation). Experimental results are shown in Table \ref{soft-table}. In all experiments, our data distillation method NACD consistently yielded improvements on both test datasets when Flan v2 and all four datasets were used as target datasets. Specifically, when Flan v2 was utilized as the target dataset, our method achieved a notable increase of 2.7\% over the original LESS results in the case of ARC-Easy.

\begin{table}
  \caption{Results of the instruction tuning tasks in MMLU and ARC-Easy for the embedding layer consistency model. The text distillation process is conducted in OPT-1.3B and so is the base model of instruction tuning. \textbf{Bold} numbers denote that our proposed method outperforms the baselines. \underline{Underlined} numbers denote the best methods with 5\% data training.}
  \label{soft-table}
  \centering
  \begin{tabular}{lllcccc}
    \toprule
    \multirow{2}{*}{Base Model}&\multirow{2}{*}{Method}&\multirow{2}{*}{Data \%}&\multirow{2}{*}{Full dataset}&\multirow{2}{*}{\# Full dataset}&MMLU&ARC-Easy\\
    \cmidrule(r){6-7}
    &&&&&Accuracy&Accuracy\\
    \midrule
    \multirow{7}{*}{OPT-1.3B}&baseline&0\%&-&-&29.6&51.1\\
    &FULL&100\%&Flan v2&100000&30.2&54.9\\
    &LESS&5\%&Flan v2&100000&29.4&49.1\\
    &NACD&5\%&Flan v2&100000&\textbf{29.8}&\textbf{51.8}\\
    &FULL&100\%&$All^*$&270679&30.0&54.0\\
    &LESS&5\%&$All^*$&270679&30.0&52.4\\
    &NACD&5\%&$All^*$&270679&\textbf{\underline{30.1}}&\textbf{\underline{52.7}}\\
    \bottomrule
  \end{tabular}
\end{table}

\textbf{For the embedding layer inconsistency model:} As a data distillation method, the transferability of synthetic data across models has always been a crucial criterion to consider. There have been a series of synthetic text representation distillation methods \cite{li2021data, Sucholutsky_2021, maekawa-etal-2023-dataset, sahni2023exploring}, but they were limited to being used only within the distilled models. For models that do not share an embedding space (e.g., Llama-2-7B), the synthetic prompt data cannot be directly concatenated. As discussed in section \ref{sec_itandeval}, when performing instruction tuning in these models, we first find the nearest token IDs for all prompt embeddings and then convert them to texts and Llama token IDs so that the new Llama token IDs can be concatenated with the raw text data.

The experimental results of the embedding layer inconsistency model are shown in Table \ref{cross-table}. Referring to \cite{xia2024less}, we also introduced non-language model-based methods such as BM25 \cite{robertson2009probabilistic} and DSIR \cite{xie2023data} for comparison and conducted the prompt data distillation in all four text dataset. As we can see, our method can significantly improves the results of instruction tuning. And the approaches of integrating knowledge from large models (LESS and ours) have shown clear superiority over other data selection methods (BM25 and DSIR).

Overall, our method achieved the best performance under the condition of training with only 5\% of data for both types of the embedding layer consistency model and the embedding layer inconsistency model. Our data distillation approach resulted in significant improvements for the instruction tuning task when applied to the selected small text dataset.

\begin{table}
  \caption{Results of the instruction tuning tasks in MMLU and ARC-Easy for the embedding layer inconsistency model. The text distillation process is conducted in OPT-1.3B, where the base model of instruction tuning is set to Llama-7B. \textbf{Bold} numbers denote that our proposed method outperforms the baselines. \underline{Underlined} numbers denote the best methods with 5\% data training. '\#' denotes the results from \cite{xia2024less}}
  \label{cross-table}
  \centering
  \begin{tabular}{lllcc}
    \toprule
    \multirow{2}{*}{Base Model}&\multirow{2}{*}{Method}&\multirow{2}{*}{Data \%}&MMLU&ARC-Easy\\
    \cmidrule(r){4-5}
    &&&Accuracy&Accuracy\\
    \midrule
    \multirow{7}{*}{Llama-2-7B}&baseline&0\%&46.0&66.8\\
    &FULL&100\%&51.1&79.0\\
    &BM25&5\%&$47.6^\#$&-\\
    &DSIR&5\%&$46.1^\#$&-\\
    &LESS&5\%&49.7&75.6\\
    &NACD&5\%&\textbf{\underline{49.8}}&\textbf{\underline{76.1}}\\
    \bottomrule
  \end{tabular}
\end{table}

\section{Analysis}
\subsection{Influence of the selected data and the statistical significance}  \label{sec_select}  

\begin{wraptable}{r}{7cm}
  \caption{Results of NACD on random samples}
  \label{randtable}
  \centering
  \begin{tabular}{lllc}
    \toprule
    \multirow{2}{*}{Base Model}&\multirow{2}{*}{Method}&\multirow{2}{*}{Data \%}&ARC-Easy\\
    \cmidrule(r){4-4}
    &&&Accuracy\\
    \midrule
    \multirow{4}{*}{OPT-1.3B}&baseline&0\%&51.1\\
    &FULL&100\%&54.0\\
    &RAND&5\%&52.7(0.2)\\
    &NACD&5\%&\textbf{\underline{53.6(0.4)}}\\
    \bottomrule
  \end{tabular}
\end{wraptable}

To demonstrate the generalizability of our data distillation method and examine the impact of selected data, we replaced LESS with random sampling for data selection in this section. Besides this, to prove the statistical significance of our method, we have added experimental results of randomly selected data. Following the setting in \cite{xia2024less}, we ran the experiment three times with different seeds and provided the error bars.The OPT 1.3B experimental results when full dataset is $All^*$ are shown in Table \ref{randtable}. For randomly selected samples, NACD also brings improvements to ARC-Easy experimental results, which demonstrating our approach has few dependencies on specific data selection methods and opening up a new perspective for data compression LLM training.


\subsection{Ablation studies}  
To validate the role of the regularization loss proposed in our Eq. \ref{reg_eq}, we conducted experiments using OPT-1.3B as an example. During the data distillation process, we experimented with the effects of incorporating and omitting this loss term. According to the results shown in Table \ref{regtable}, it is evident that the introduction of this loss function significantly enhances the effectiveness of the data distillation method. From a broader perspective, enabling the learned pseudo embeddings to have a sharp distribution over the token vocabulary seems to provide synthetic data with more distinct semantics, thereby increasing the robustness of the results.

\begin{table}
  \caption{Ablation study about the regularization loss proposed in Eq. \ref{reg_eq}}
  \label{regtable}
  \centering
  \begin{tabular}{llllcc}
    \toprule
    \multirow{2}{*}{Base Model}&\multirow{2}{*}{Method}&\multirow{2}{*}{Data \%}&\multirow{2}{*}{Full dataset}&MMLU&ARC-Easy\\
    \cmidrule(r){5-6}
    &&&&Accuracy&Accuracy\\
    \midrule
    \multirow{2}{*}{OPT 1.3B}&with reg&5\%&Flan v2&29.8&51.8\\
    &w.o. reg&5\%&Flan v2&29.5&51.7\\
    \midrule
    \multirow{2}{*}{OPT 1.3B}&with reg&5\%&$All^*$&30.1&52.7\\
    &w.o. reg&5\%&$All^*$&29.8&52.4\\
    \bottomrule
  \end{tabular}
\end{table}

The compression ratio was a important factor in NACD, in order to verify the stability of NACD in different scenarios, under the setting of "ARC-Easy+$All^*$", we added experiments with larger compression rates of 10\% and smaller compression rates of 0.02\%. As shown in Table \ref{ablation_table_ratio}, at different compression rates, NACD consistently shows improvement, further confirming the effectiveness of our method. Additionlly, we have added an ablation study on prompt length k and regularization coefficient $\beta$ in Table \ref{ablation_table_beta} . From the experimental results, it can be observed that, firstly, for different values of beta (100, 50, and 20), NACD shows stable performance. While we also attempted to increase k to 5, and even NACD exhibited better performance. However, when we increased k to 10, NACD started showing disastrous performance. We believe that the increase of the learning space has led to an excessive level of difficulty in learning.  Therefore, in NACD, there exists a balance between the search space of data and the learning difficulty.

\begin{table}
  \caption{Different compression ratios experimental results of the instruction tuning tasks in ARC-Easy. \textbf{Bold} numbers denote that our proposed method outperforms the baselines. NACD shows stable improvement in LESS under different compression ratios.}
  \label{ablation_table_ratio}
  \centering
  \begin{tabular}{llccc}
    \toprule
   {Base Model}&{Method}&{Full dataset}&{Data \%}&Accuracy\\
    \midrule
    \multirow{8}{*}{OPT-1.3B}&baseline&$All^*$&0\%&51.1\\
    &FULL&$All^*$&100\%&54.0\\
    &LESS&$All^*$&10\%&53.2\\
    &LESS+NACD&$All^*$&10\%&\textbf{53.8}\\
    &LESS&$All^*$&0.02\%&51.1\\
    &LESS+NACD&$All^*$&0.02\%&\textbf{51.6}\\
    \bottomrule
  \end{tabular}
\end{table}

\begin{table}
  \caption{Ablation studies of the instruction tuning tasks in ARC-Easy. $k$ denotes the prompt token length and $\beta$ denotes the weight factor in Equation \ref{reg_eq}. To specifically validate the effectiveness of the NACD method on other source data, we replaced the target full dataset with Dolly.}
  \label{ablation_table_beta}
  \centering
  \begin{tabular}{llccccc}
    \toprule
   {Base Model}&{Method}&{Full dataset}&$\beta$&$k$&{Data \%}&Accuracy\\
    \midrule
    \multirow{7}{*}{OPT-1.3B}&baseline&Dolly&-&-&0\%&51.1\\
    &FULL&Dolly&-&-&100\%&54.0\\
    &LESS&Dolly&-&-&5\%&50.8\\
    &LESS+NACD&Dolly&100&5&5\%&52.0\\
    &LESS+NACD&Dolly&100&2&5\%&51.0\\
    &LESS+NACD&Dolly&50&2&5\%&51.2\\
    &LESS+NACD&Dolly&20&2&5\%&51.3\\
    \bottomrule
  \end{tabular}
\end{table}

\subsection{The computation cost}
Table \ref{computationtable} describes the computation complexity and the actual time of NACD. Taking the OPT-1.3B experiment of ARC-Easy in Flan-v2 as an example, we can see the main computation time costs in expert trajectory and data distillation steps. Noteworthy is the fact that in practical implementation, once synthesized data is generated through the initial two steps, it is important to note that the distilled dataset obtained at once can be used multiple times for model training, and even for instruction learning tasks in different language models (the embedding layer inconsistent model in Section \ref{mainresultsec}). This one-time effort in data synthesis proves to be a cost-effective approach, as the data distillation paradigm significantly reduces data volume, leading to substantial savings in subsequent instruction

\begin{table}
  \caption{The computation complexity and the actual time of NACD. All experiments in this paper are conducted in 8 Nvidia A800 GPUS}
  \label{computationtable}
  \centering
  \begin{tabular}{cccc}
    \toprule
    &Expert trajectory extraction&data distillation&instruction tuning\\
    \midrule
    Complexity&$\mathcal{O}(|\{\tau_{i}^{*}\}| \cdot M \cdot Epoch_{step1})$&$\mathcal{O}(S \cdot N)$&$\mathcal{O}(M\cdot p_r \cdot Epoch_{step3})$\\
    Actual time& $\sim 36 hours$&$\sim 28 hours$&$\sim 15 minutes$\\
    \bottomrule
  \end{tabular}
\end{table}

\subsection{Visualization of the synthesized data}

The synthetic data obtained from NACD distillation is pseudo embedding. In order to observe the semantics of the synthesized data, we take the distillation experiment with Flan v2 on ARC-Easy as an example, select two samples for analysis and decode the nearest pseudo-tokens as shown in the Figure \ref{Fig.examples}. It can be observed that the synthesized prompt data tends to provide hints towards the correct answers, such as "disaster" and "suspected" for negative reviews recognition. Additionlly


\begin{figure}[htbp]
	\centering
    \begin{center}
    \fcolorbox{black}{gray!10}{\parbox{.9\linewidth}{
    \centerline{\bf Example 1} 
    {\bf Decoded Pseudo Tokens:}[disaster, </s>]\\
    {\bf Original Data:}\\
    \textit{User}: Pick from:
     A). Yes
     B). No
    Q: Title: bad production Review: the production of this dvd was very low, could hardly hear what the comedians where saying on stage without having a very high volume on my sound system.great comedians, but the quality of the dvd pulled down the enjoyment of it all unfortunantly. Is this product review negative?
    A:\\
    \textit{Assitant}: A).\\
    \centerline{\bf Example 2} 
    {\bf Decoded Pseudo Tokens:} [ask, suspected]\\
    {\bf Original Data:}\\
    \textit{User}: Q: Title: Unreliable Review: Difficult to read the raised plastic numbers on the dials. Not user friendly to set. The 5-minute watering interval settings are not accurate. Recommend a digital model instead. Is this product review negative?
    
    pick from the following.
     -- Yes
     -- No
    The answer is:
    A:\\
    \textit{Assitant}: Yes
    }}
    \end{center}
    \caption{Samples to show the effects of distilled pseudo tokens.}
    \label{Fig.examples}
\end{figure}

\section{Conclusion}
In this paper, we discuss the text data distillation in text generation tasks, i.e., instruction tuning. We combine prompt tuning and trajectory matching to synthesize an efficient prompt token dataset which can approximate or even surpass the performance of the full dataset when training a language model based on a selected subset. Experimental results prove the effectiveness of our proposed method. Our synthetic data has yielded a maximum benefit of 2.7\% to the original data selection result.

In the future, we will try to extend our experiments to different NLP tasks and LLMs. We can also apply our method in more complex scenarios, such as synthesizing multimodal data.

\bibliographystyle{unsrt}
\bibliography{main}

\end{document}